\documentclass[11pt]{article}

\usepackage[final]{acl}

\usepackage{comment}
\usepackage{times}
\usepackage{latexsym}
\usepackage[T1]{fontenc}

\usepackage{microtype}
\usepackage{inconsolata}
\usepackage{graphicx}
\usepackage{hyperref}
\usepackage{scalerel}
\usepackage{tikz} 

\usepackage{booktabs}
\usepackage{longtable}
\usepackage{array}
\usepackage{ragged2e}

\usepackage{multirow}
\usepackage{booktabs}
\usepackage[table]{xcolor}
\usepackage{xcolor}
\usepackage{caption}
\usepackage{subcaption}
\usepackage{threeparttable}
\usepackage{adjustbox}
\usepackage{amsmath}
\definecolor{bestcell}{HTML}{C6E5D9}   
\definecolor{secondcell}{HTML}{DCE6F2} 
\definecolor{thirdcell}{HTML}{F2A191}  
\definecolor{orcidlogocol}{HTML}{A6CE39}
\newcommand{\orcidicon}[1]{%
    \href{https://orcid.org/#1}{%
        \begin{tikzpicture}[baseline=-0.45ex]
            \draw[orcidlogocol, fill=orcidlogocol] (0,0) circle [radius=0.16] node[white] {{\fontfamily{qhv}\selectfont \tiny \textbf{iD}}};
        \end{tikzpicture}
    }%
}

\title{Wisdom in Unity: The Role of Multilingual Training in Figurative Language Identification in Proverbs}

\author{
\textbf{Rama Alomair} \orcidicon{0009-0005-0476-6454}
\quad
\textbf{Remas Alsubaie} \orcidicon{0009-0008-7468-8860}
\quad
\textbf{Walaa Saifalislam} \orcidicon{0009-0002-4107-6821}
\quad
\textbf{Rima Alsonbul} \orcidicon{0009-0009-4254-5707}
\\[3pt]
\textbf{Mona Alnajjar} \orcidicon{0009-0009-1871-032X}
\quad
\textbf{Razan Aldossari} \orcidicon{0009-0004-6755-8954}
\quad
\textbf{Haya Alibrahim} \orcidicon{0009-0009-1805-1010}
\quad
\textbf{Abeer Aldayel} \orcidicon{0000-0002-0843-0123}
\\[3pt]
College of Computer and Information Sciences, King Saud University
\\[2pt]
{\fontsize{7}{8}\selectfont
\texttt{\{rama.alomair01,monasalehalnajjar,remasalsubaie207,rikhmoso1425\}@gmail.com}
}
\\[-1pt]
{\fontsize{7}{8}\selectfont
\texttt{aabeer@ksu.edu.sa}
}
}
\begin{document}
\maketitle

\begin{abstract}
Although multilingual approaches to figurative language identification are not new, the shift beyond language-homogeneous training data requires a clearer understanding of the contribution of translated multilingual supervision. We examine this question using 742 proverb concepts across 6,787 translated instances for seven languages. We evaluate five models including multilingual encoders and instruction-tuned LLMs through progressively increasing levels of multilingual supervision. Moreover, we introduce multidimensional annotation framework for proverbs that characterizes proverbs through four complementary figurative forms: Metaphorical, Moral/Advisory, Cause–Effect, and Culture-Specific.

Our findings show that overall, adding multilingual training data beyond 50\% provides only limited additional improvement, although the best supervision level varies across models and languages. Also, we show that combining diverse figurative forms yields the strongest overall performance. A notable finding is that the least frequent figurative form ``culture-specific'' exhibits the largest performance gains under multilingual supervision. Furthermore, the moral/advisory and culture-specific forms of proverb contribute more to instruct tuning LLM overall figurative identification performance. 
These findings motivate multilingual figurative identification to move beyond metaphor-centric taxonomies toward concept-level multidimensional frameworks that explicitly model complementary forms of figurative meanings that are context representative.

\end{abstract}

\section{Introduction}

The interpretation of figurative language, particularly proverbs, represents a pinnacle of linguistic nuance that remains a bottleneck for modern Natural Language Processing \citep{ghosh-etal-2023-figbench}. Unlike literal text, the semantic essence of a proverb is often ``non-compositional'' meaning its global significance cannot be derived solely from the individual meanings of its constituent words~\citep{Hrisztova-Gotthardt2015-vi}. 
As proverbs are rich carriers of cultural knowledge, they have become a common test bed for multilingual figurative language understanding~\citep{Liu2024-ab}. Prior work has treated figurativeness as a linguistic phenomenon by focusing on rhetorical devices such as metaphors~\citep{Jang2017-nz}, similes~\citep{Niculae2014-ln}, or idioms~\citep{Oh2026-rm}. Instead, in this study we provide conceptually based multidimensional framework that models complementary forms of figurative meaning in order to enable a finer-grained analysis of how figurative meaning is expressed and transferred across multilingual proverb translations.

Mainly, we introduce a fine-grained diagnostic annotations for analyzing complementary dimensions of figurative forms in proverbs. Using multilingual proverbs as a testbed, we show how four forms of figurative language are affected by different levels of multilingual supervision. According to Relevance Theory~\citep{Sperber1986-oq}, communication is inferential rather than purely code-based. In our case, proverbs comprehension may involve different kinds of inferential knowledge. Thus, we provide four forms as a study’s operationalization of interpretive dimensions of figurative forms in proverbs as: metaphorical meaning expressed through imagery, analogy, or symbolic entities; moral/advisory meaning that conveys guidance, warning, or a lesson; cause–effect meaning that links actions with their outcomes; and culture-specific symbolic meaning that requires cultural, religious, historical, or local knowledge for full interpretation.  Then we examine the multilingual training effect on figurative identification, leveraging a multilingual proverb dataset and focusing the evaluation on translations into seven diverse target languages: Arabic, English, French, German, Russian, Japanese, and Spanish. 

Accordingly, our study addresses the following research questions:

\begin{itemize}
\item \textbf{RQ1.} \textit{To what extent does progressively incorporating aligned multilingual proverb translations improve multilingual figurative language identification under limited supervision?}
We evaluate whether increasing the proportion of translated training data (0\%, 10\%, 50\%, and 100\%) improves multilingual figurative identification across different model and languages.
\item \textbf{RQ2.} \textit{ Which forms of figurative forms in proverbs are associated with overall figurative identification performance, and which forms benefit most from increasing multilingual supervision?}
\item \textbf{RQ3.} \textit{Which forms of figurative language are preserved across translated proverbs?}
\end{itemize}

To answer these questions, we complement binary figurative classification with a human-annotated semantic analysis of proverb structure, examining whether different figurative forms benefit equally from multilingual transfer or exhibit different transfer characteristics. The contributions of this paper are:

\begin{itemize}
\item We evaluate the effect of progressively increasing multilingual supervision (0\%, 10\%, 50\%, and 100\%) on multilingual proverbs figurative identification across 7 languages using different models.
\item We introduce a fine-grained annotation framework that provides an additional diagnostic layer for analyzing how different dimensions of figurative forms transfer across languages in proverbs.
\item We show that multilingual supervision benefits of figurative forms unevenly and providing insights on which forms of figurative knowledge are most transferable.
\end{itemize}

\section{Related Work}

\paragraph{Figurative Detection in Multilingual Setting}
Recent research on figurative language detection explore with various methods and demonstrate the effectivness of different classification methods on figurative identification \citep{chakrabarty-etal-2022-flute}.
A prominent example of this trend is SemEval-2022 Task 2 \citep{madabushi-etal-2022-semeval}, which framed multilingual idiomaticity detection as binary classification in context and sentence-level representation.
This modeling trend is reinforced by other works such as \citet{yamaguchi-etal-2022-hitachi} built their  around a suite of multilingual pretrained language models, specifically mBERT, XLM-R, InfoXLM, XLM-Align, and RemBERT, and compared their effectiveness for idiomaticity classification. Beyond idiomatic detection, recent benchmark work has shown that figurative language remains a difficult challenge even for strong neural models. \citet{chakrabarty-etal-2022-flute} introduced FLUTE, a dataset of 9,000 figurative natural language inference instances with human-written explanations, covering sarcasm, simile, metaphor, and idioms. Also, the work by \citet{lai-etal-2023-mmfld} introduced MMFLD, a benchmark spanning seven languages and multiple figures of speech. Another work by~\citet{Lai2023-as} provides a framework for figurative language detection using template-based prompt learning without requiring language-specific modules. The work by ~\citet{Hulsing2024-eg} compares neural and non-neural cross-lingual models for English as the source language and Russian, German, and Latin as target languages, showing the best performance gained by their neural cross-lingual adapter architecture. Focusing on cultural representation the work by ~\citep{Kabra2023-ib} they introduced MABL as a culturally grounded figurative-language inference dataset across seven diverse languages demonstrating that cultural concept shifts substantially challenge multilingual models beyond purely cross-lingual transfer.

\paragraph{Figurative Detection in Proverbs} Despite these advances, proverb-focused figurative language identification remains less studied than idiom detection or broader figurative benchmarks, particularly in multilingual settings where cultural knowledge and indirect meaning play a central role. Most of the work focused on monolingual proverbs such as FFE-HALLU, Persian~\citep{Hosseini2026-mk} and Jawaher, Arabic~\citep{Magdy2025-pu}. Another line of work examines downstream tasks performance, such as translation effect ~\citep{Wang2025-at}, sentiment analysis~\citep{Alsiyat2020-wb} , and narrative analysis~\citep{Chakrabarty2022-ih,Alshaalan2026-zd}. 

Proverbs are often short, implicit, and highly culture-dependent, which makes their figurative forms difficult to infer through literal interpretation alone. This makes proverb identification a particularly challenging test case for multilingual supervision. Thus, most of the previous work focuses on identifying established figures of speech, such as metaphors~\citep{Ozbal2016-dy,Goren2024-gw}, similes~\citep{Khoshtab2025-as}, idioms~\citep{Almheiri2026-zv}, and hyperbole~\citep{Badathala2023-eu}. For example,~\citet{Ozbal2016-dy} formulates proverb analysis as a word-level metaphor recognition task to identify metaphorically used words within proverb expressions. In contrast, our study characterizes proverbs at the concept level through a multidimensional annotation framework that captures not only metaphorical realization, but also cause and effect relations, moral or advisory functions, and dependence on culture-specific knowledge. These categories describe complementary dimensions of how a proverb constructs and communicates its figurative meanings rather than treating figurative as a single linguistic device.

\begin{table}[t]
\centering
\scriptsize
\setlength{\tabcolsep}{4.5pt}
\renewcommand{\arraystretch}{1.08}
\begin{tabular}{lrrrr}
\toprule
\textbf{Lang} &
\textbf{Unique Concepts} &
\textbf{Instances} &
\textbf{Lit} &
\textbf{Fig} \\
\midrule
Arabic   & 149 & 226  & 134 & 92  \\
English  & 651 & 1,345 & 689 & 656 \\
French   & 580 & 1,152 & 598 & 554 \\
German   & 578 & 1,140 & 577 & 563 \\
Japanese & 413 & 835  & 418 & 417 \\
Russian  & 567 & 1,095 & 567 & 528 \\
Spanish  & 514 & 994  & 502 & 492 \\
\midrule
\textbf{Overall} &
\textbf{742} &
\textbf{6,787} &
\textbf{3,485} &
\textbf{3,302} \\
\bottomrule
\end{tabular}
\caption{Distribution of proverb concepts and multilingual instances across the seven target languages across Literal (Lit) and Figurative (Fig). Proverb concepts may occur in multiple languages, the language-level concept counts do not sum to the overall total of 742 unique concepts.}
\label{tab:language-dataset-distribution}
\end{table}
\section{Dataset}

To investigate the effectiveness of multilingual supervision for figurative language identification, we constructed a multilingual ground-truth dataset from existing figurative language resources\footnote{The dataset and annotations can be accessed https://osf.io/vc4mt}. We enrich the labeling for the rest of the languages by incorporating silver labels for unlabeled multilingual proverbs using a weak supervision model.

\begin{table}[t]
\centering
\small
\setlength{\tabcolsep}{5pt}
\renewcommand{\arraystretch}{1.1}

\begin{tabular}{lcc}
\toprule
\textbf{Figurative form} &
\textbf{GroundTruth} &
\textbf{All instances} \\
\midrule
Metaphorical      & 55.4\% & 47.0\% \\
Literal (None)    & 42.2\% & 30.7\% \\
Cause-Effect     & 28.9\% & 45.2\% \\
Moral/Advisory    & 24.1\% & 42.9\% \\
Culture-Specific  & 6.0\%  & 13.4\% \\
\bottomrule
\end{tabular}

\caption{Distribution of the fine-grained figurative forms in the manually annotated ground-truth subset of 83 proverb concepts and the weakly labeled  dataset of 6,787 instances. Percentages are non-exclusive as the annotation schema is multi-label. }
\label{tab:label-distribution}
\end{table}


\begin{table}[t]
\centering
\scriptsize
\begin{tabular}{lrrrr}
\toprule
Split & Concepts & Rows & Literal & Figurative\\
\midrule
Train      & 518 & 4777 & 2518 & 2259\\
Validation & 75  & 728  & 306  & 422\\
Test       & 149 & 1282 & 661  & 621\\
\midrule
Total      & 742 & 6787 & 3485 & 3302\\
\bottomrule
\end{tabular}
\caption{Binary label distribution. The dataset is nearly balanced overall (51.35\% literal and 48.65\% figurative).}
\label{tab:data_split}
\end{table}


\begin{figure}[t]
    \centering
    \includegraphics[width=0.70\columnwidth]{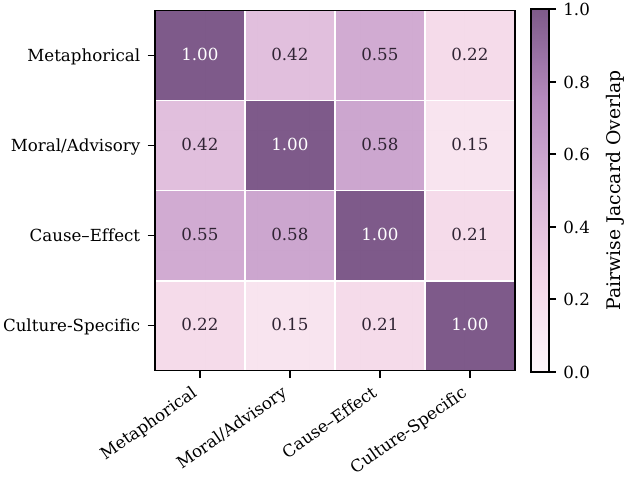}
    \caption{Pairwise Jaccard similarity between the four fine-grained figurative forms computed at the unique proverb-concept level. Higher values: two labels are assigned to a larger proportion of the same concepts, lower values labels show distinct figurative properties. }
    \label{fig:finegrained_jaccard}
\end{figure}

\subsection{Proverbs Collection}

Our experiments are based on a multilingual proverb translation dataset \citep{multilingual-proverb-dataset,Alshaalan2026-zd} containing 25,300 aligned proverb pairs spanning multiple languages and cultural traditions. Each record consists of a source proverb and one of its translated counterparts together with their language identifiers and alignment metadata. To reduce translation redundancy, we retained at most three translated variants for each source proverb-target language pair, following prior multilingual data-selection practices that limit repeated near-duplicate translations while preserving linguistic diversity~\citep{chuang-etal-2023-so-many}.

After filtering, the final dataset used in this study contains  742 unique proverb concepts represented by 6,787 multilingual instances, Table~\ref{tab:language-dataset-distribution}. Concept-level counts are used for the annotation-distribution and label-overlap analyses. Each proverb concept may be represented in several languages and by up to three translated variants per source-concept--target-language pair.
As shown in table~\ref{tab:data_split}, the dataset is split into 4,777 training instances (518 concepts), 728 validation instances (75 concepts), and 1,282 test instances (149 concepts), yielding an overall dataset of 3,302 figurative and 3,485 literal examples. The binary label distribution is therefore nearly balanced (48.65\% figurative and \ 51.35\% literal), providing a suitable benchmark for multilingual figurative language identification. 
\paragraph{Figurative and Fine-grain forms annotations} We ground our annotation framework in Relevance Theory~\citep{Sperber1986-oq}, which views proverb comprehension as an inferential process that integrates linguistic meaning with contextual and cultural knowledge. Under this perspective, proverb interpretation may rely on different dominant inferential mechanisms, including metaphorical mapping, causes and effect, and culturally shared symbolic knowledge. Guided by this framework, we annotate each proverb using four complementary forms of figurative meanings. They are not all figurative forms in the linguistic sense. Rather, they are different ways figurative meaning is realized or interpreted. These forms are: \textit{Metaphorical}:conveys meaning indirectly through imagery or analogy, \textit{Moral/Advisory}:conveys guidance, warning or ethical judgment, \textit{Cause–Effect}:expresses a relation between an action and its consequence, and \textit{Culture-Specific}:requires cultural, religious, or locally shared knowledge for interpretation. These forms of fine grain figurative meanings are multilabels where one instance might have more than one type at once. Three of the authors acted as annotator for ground-truth labeling with 2 annotators per source ID, across 83 source ID. Then, the third annotator review and resolved any conflicted labels based on the annotation guideline provided to the annotators( guideline). The binary figurative-versus-literal classification achieved a Cohen's $\kappa$ of 0.626 (81.9\% raw agreement), indicating substantial agreement. For the multilabel figurative properties, Cohen's $\kappa$ ranged from 0.423 (Cause and Effect) to 0.622 (None/Literal), with the highest agreement for Culture-Specific Symbolic (92.8\%) (detailed annotation process reported in Appendix~\ref{app:annotationValidation}).

\paragraph{Weak Supervision Labeling across languages}
 \textbf{(a) Binary Figurative labeling.} To build a reliable ground-truth dataset that we will use fore weak-supervision to the rest of out 6K dataset, we further used three figurative language datasets were combined: MMFLD \citep{lai-etal-2023-mmfld}, MAPS \cite{liu-etal-2023-maps}, and MEMPHIS \citep{memphis-dataset}. MMFLD contributed sentences across English, German, Spanish, Italian, and Chinese, covering idiom, hyperbole, and simile categories. MAPS added short proverbs across six languages (English, German, Russian, Indonesian, Chinese, and Bengali), exposing the model to short, implicit figurative expressions similar in style to the target proverbs. MEMPHIS contributed Arabic idiom examples, directly addressing the Arabic coverage gap. All datasets were unified into a binary classification format, where label = 0 represents literal expressions and label = 1 represents figurative expressions. With net dataset consist of about 325,298 instances, including 260,238 training, 32,530 validation, and 32,530 test examples. Overall, the dataset contains 139,002 figurative (42.7\%). Then after models comparison we use XLM-RoBerta Base for weak-supervision binary labeling (detail experiment are shown in Appendix~\ref{app:annotationValidation}, Table~\ref{tab:weak-supervision-mcnemar}).

\textbf{(b) Multi-label Figurative forms.} We used E5 embeddings with Logistic Regression as the weak-supervision model for our multilabeling fine-grained labels after comparing this model with multiple setting and compare Jaccard overlap between labels (detail experiment at Appendix~\ref{app:annotationValidation}, Table~\ref{tab:finegrained_results}). This model provided the best trade-off between performance and efficiency for large-scale annotation. Moreover, Figure~\ref{fig:finegrained_jaccard} shows the pairwise Jaccard similarity among the fine-grained labels. Cause–Effect has the strongest overlap with both Moral/Advisory (0.58) and Metaphorical (0.55), whereas Culture-Specific exhibits consistently lower similarity (0.15–0.22), suggesting that it captures a more distinct meaning of figurative forms. Importantly, the moderate off-diagonal Jaccard similarities indicate that the labels are related but not interchangeable, supporting the use of separate figurative forms in the annotation framework.
\begin{figure*}[t]
    \centering
    \begin{minipage}[t]{0.49\textwidth}
        \centering
        \includegraphics[
            width=\linewidth
        ]{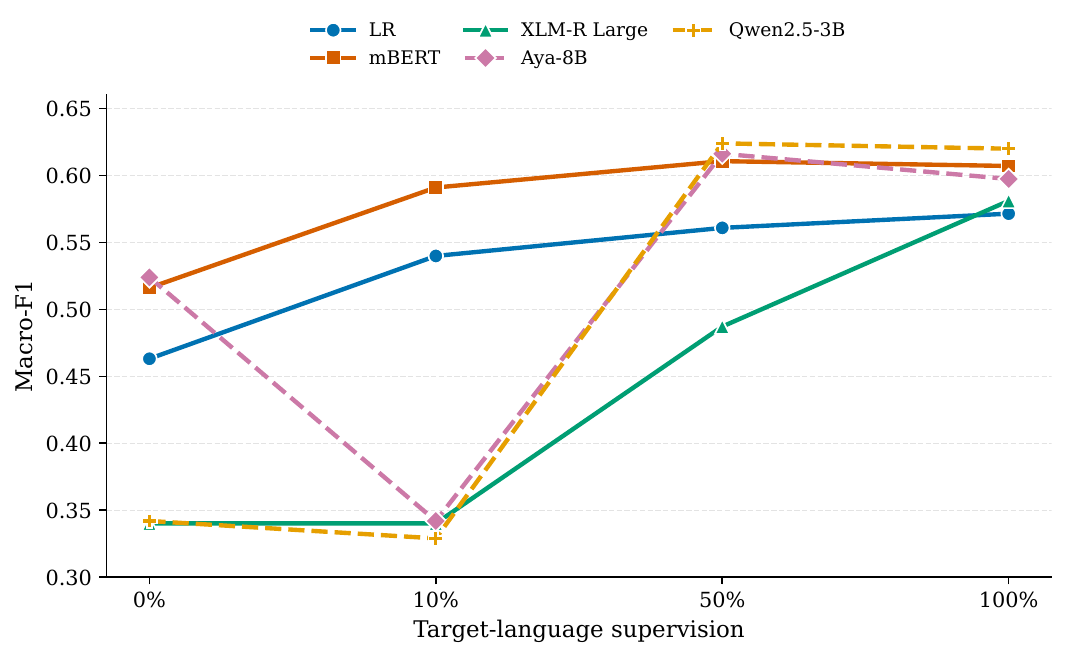}
        \captionof{figure}{
        Overall Macro-F1 across increasing levels of
        multilingual supervision.
         }
        \label{fig:macro-f1-supervision}
    \end{minipage}
    \hfill
    \begin{minipage}[t]{0.49\textwidth}
        \centering
        \includegraphics[
            width=\linewidth
        ]{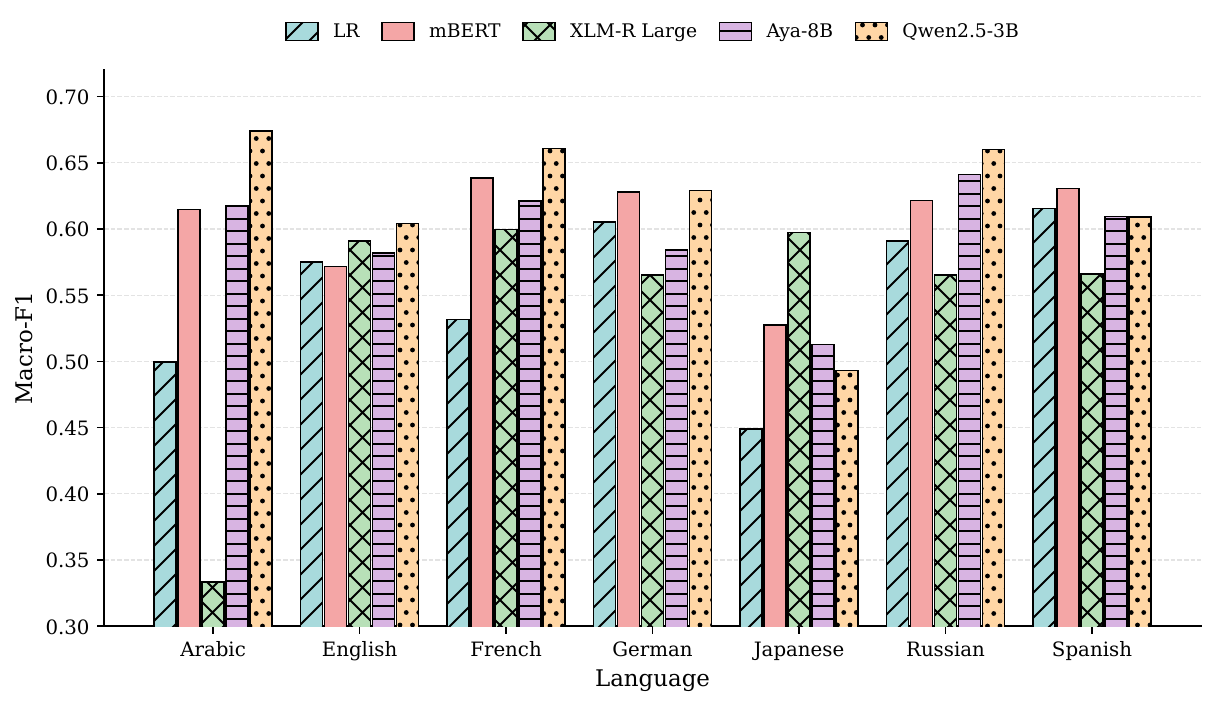}
        \captionof{figure}{
        Per-language Macro-F1 at 100\% multilingual supervision. 
        }
        \label{fig:macro-f1-language-100}
    \end{minipage}

\end{figure*}
Overall, Table~\ref{tab:label-distribution} summarizes the distribution of the fine-grained figurative forms in both the manually annotated ground-truth subset and the complete weakly labeled multilingual dataset. In both datasets, Metaphorical is the most prevalent figurative form, whereas Culture-Specific is consistently the least frequent. Although the full dataset contains relatively more Cause–Effect and Moral/Advisory labels and fewer Literal (None) instances than the annotated subset, the overall distribution remains consistent. This suggests that the weak-supervision process preserves the general characteristics of the manually annotated seed while scaling the annotations to the full multilingual dataset. Moreover, to evaluate silver-label quality, we manually validated 149 held-out proverb concepts against the weakly supervised annotations. The comparison shows a moderate disagreement as shown by  Hamming error rate of 0.330 and a moderate sample-level Jaccard similarity of 0.439 (detailed per-label results are reported in Appendix~\ref{app:annotationValidation}).

\section{Methodology}
We conduct two experiments to evaluate the effect of multilingual transfer on figurative language identification in proverb (binary detection) (i) progressive multilingual supervision (Section~\ref{sec:progressive-supervision}) and (ii) fine-grained instance-selection ablation (Section~\ref{sec:ablation}). To examine whether the observed trends generalize across model families, we evaluate five representative architectures: a Logistic Regression baseline, Multilingual BERT (mBERT)~\citep{Devlin2018-qn}, XLM-RoBERTa Large~\citep{Conneau2019-rx}, Aya Expanse 8B (23 languages)~\citep{Dang2024-ay}, and Qwen2.5-3B (29+ languages)~\citep{Qwen-Team2024-ih} (detail models configuration in Appendix~\ref{app:models_settings}). Together, these models cover classical machine learning, multilingual encoder-based transformers, and multilingual instruction-tuned LLMs, enabling a comprehensive analysis of multilingual supervision and language-specific behavior across the seven target languages. In these experiments, we followed a unified evaluation setting where all models are evaluated on the same fixed validation and test sets. Binary figurative-identification performance is measured using Macro-F1 as the primary metric. Then, Fine-grained analyses are performed by evaluating the trained binary classifiers on test subsets associated with each figurative form, while the underlying training and prediction tasks remain unchanged.

\subsection{Progressive multilingual supervision.}
\label{sec:progressive-supervision}
We evaluate progressively increasing levels of multilingual supervision while keeping the source-language training data fixed. The source-language portion was always included in training, while the proportion of multilingual examples added on top was systematically varied across four supervision levels. The 0\% setting uses only source-language training instances, whereas the 10\%, 50\%, and 100\% settings progressively add 10\%, 50\%, and 100\% of the translated multilingual training instances, respectively. Throughout the experiments, the validation and test sets remain unchanged, allowing performance differences to be attributed exclusively to the amount of multilingual training data.

\subsection{Fine-grained instance training ablation.} 
\label{sec:ablation}
To isolate the effect of different figurative forms on the binary figurative detection, we conducted a controlled instance selection ablation. The binary figurative identification task remained unchanged, while fine-grained labels were used only to determine which multilingual training instances were added. For each figurative form, we randomly sampled the same number of multilingual instances (600 instances) and combined them with the complete source-language training set. This controlled setup ensures that performance differences reflect the composition of the selected figurative forms rather than differences in training set size.
\subsection{Matched English vs Multilingual Control Supervision}
\label{sec:matched-control}
We conducted two augmentation conditions to control and verify the effect of language diversity from that of increased training set size. More specifically, we start from the same source language training core then we add (1) either $k=300$ English, \textit{English condition}, or (2)$k=300$ non-english translations, \textit{Multilingual condition}. The multilingual condition is balanced across the non-English languages (50 instances per language). Both conditions are evaluated on the same held-out non-English test set and we repeat each condition using five random seeds and provide the mean Macro-F1 (Table~\ref{tab:english-vs-multilingual-matched}).
\begin{table*}[t]
\centering
\scriptsize
\setlength{\tabcolsep}{8pt}
\renewcommand{\arraystretch}{1.12}

\begin{tabular}{lccccc}
\toprule
\textbf{Training condition}
& \textbf{LR}
& \textbf{mBERT}
& \textbf{XLM-R Large} 
& \textbf{Qwen}
& \textbf{Aya} 
\\
\midrule

Source only
& \cellcolor{thirdcell}0.463 $\pm$ 0.000
& \cellcolor{thirdcell}0.524 $\pm$ 0.013
& 0.336 $\pm$ 0.007 
& 0.362 $\pm$ 0.050
& 0.453 $\pm$ 0.062
\\

+ Random target
& \cellcolor{secondcell}0.530 $\pm$ 0.017
& 0.521 $\pm$ 0.098
& \cellcolor{secondcell}0.408 $\pm$ 0.118
& 0.381 $\pm$ 0.028
& 0.341 $\pm$ 0.001
\\

+ Metaphorical
& 0.450 $\pm$ 0.010
& \cellcolor{secondcell}0.571 $\pm$ 0.019
& 0.326 $\pm$ 0.000 
& 0.383 $\pm$ 0.095
& \cellcolor{thirdcell}0.459 $\pm$ 0.115
\\

+ Moral/Advisory
& 0.418 $\pm$ 0.022
& 0.454 $\pm$ 0.110
& 0.326 $\pm$ 0.000 
& \cellcolor{secondcell}0.506 $\pm$ 0.030
& 0.326 $\pm$ 0.000
\\

+ Cause-Effect
& 0.440 $\pm$ 0.014
& 0.512 $\pm$ 0.045
& \cellcolor{thirdcell}0.364 $\pm$ 0.065 
& 0.452 $\pm$ 0.109
& 0.326 $\pm$ 0.000
\\

+ Culture-Specific
& 0.367 $\pm$ 0.000
& 0.499 $\pm$ 0.033
& 0.326 $\pm$ 0.000 
& \cellcolor{thirdcell}0.493 $\pm$ 0.147
& \cellcolor{secondcell}0.505 $\pm$ 0.046
\\

\midrule

+ All target
& \cellcolor{bestcell}\textbf{0.572 $\pm$ 0.000}
& \cellcolor{bestcell}\textbf{0.598 $\pm$ 0.011}
& \cellcolor{bestcell}\textbf{0.568 $\pm$ 0.061} 
& \cellcolor{bestcell}\textbf{0.614 $\pm$ 0.009}
& \cellcolor{bestcell}\textbf{0.594 $\pm$ 0.006}
\\

\bottomrule
\end{tabular}
\caption{Controlled instance-selection ablation on the complete test
set. We use 600 instances to isolate the effect of training size. The random-target condition controls for
gains attributable to increased training volume, whereas the
\textit{All target} condition uses the complete multilingual training
set. The highest three values in each model column are highlighted using
\colorbox{bestcell}{teal} for the first,
\colorbox{secondcell}{blue} for the second, and
\colorbox{thirdcell}{coral} for the third-highest result.}
\label{tab:controlled_instance_ablation}
\end{table*}
\section{Results}
 To answer the first research question  \textbf{RQ1}: \textit{Does progressively increasing multilingual supervision improve figurative language identification?} We first analyze the performance of each models as shown in Figure~\ref{fig:macro-f1-supervision} to examine how they respond to varying scales of multilingual supervision (from 0\% to 100\%) and how their performance fluctuates across different languages. Overall, the analysis shows that all models benefit from increased supervision progressively with a variation on the magnitude of improvement across models architectures. As shown the Encoder-based models have relatively gradual gains, whereas the instruction-tuned LLMs achieve larger improvements with small amounts of additional multilingual supervision. Across all models, performance generally stabilizes after incorporating approximately 50\% of the translated multilingual training data, with smaller gains from 50\% to 100\% supervision. By zooming deeper on the languages level as illustrated in Figure~\ref{fig:macro-f1-language-100} these improvements are consistently observed across the seven target languages with a noticeable variation of the gains by language. Qwen2.5-3B achieves the highest Macro-F1 on four of the seven languages (Arabic, French, German, and Russian), while mBERT performs best on Spanish and XLM-R Large on Japanese. Also, Japanese is the most challenging language with the lowest overall Macro-F1. Interestingly, unlike the other languages where instruction-tuned LLMs generally achieve the strongest performance, XLM-R Large performs worst for Arabic and best on Japanese, suggesting that encoder-based multilingual representations remain particularly effective for this language. 
To validate that these gains are not explained solely by the larger number
of training instances we further compare English and multilingual additions at identical size. Table~\ref{tab:english-vs-multilingual-matched} illustrates that overall multilingual augmentation outperforms English addition for all five models on average, with a mean improvement of a $ 3.3$ Macro F1 increase. The improvement of multilingual augmentation is largest for Logistic Regression and mBERT, whereas XLM-R  has a small difference whose confidence
interval includes zero.

\begin{figure}[t]
    \centering
    \includegraphics[
        width=0.50\textwidth
    ]{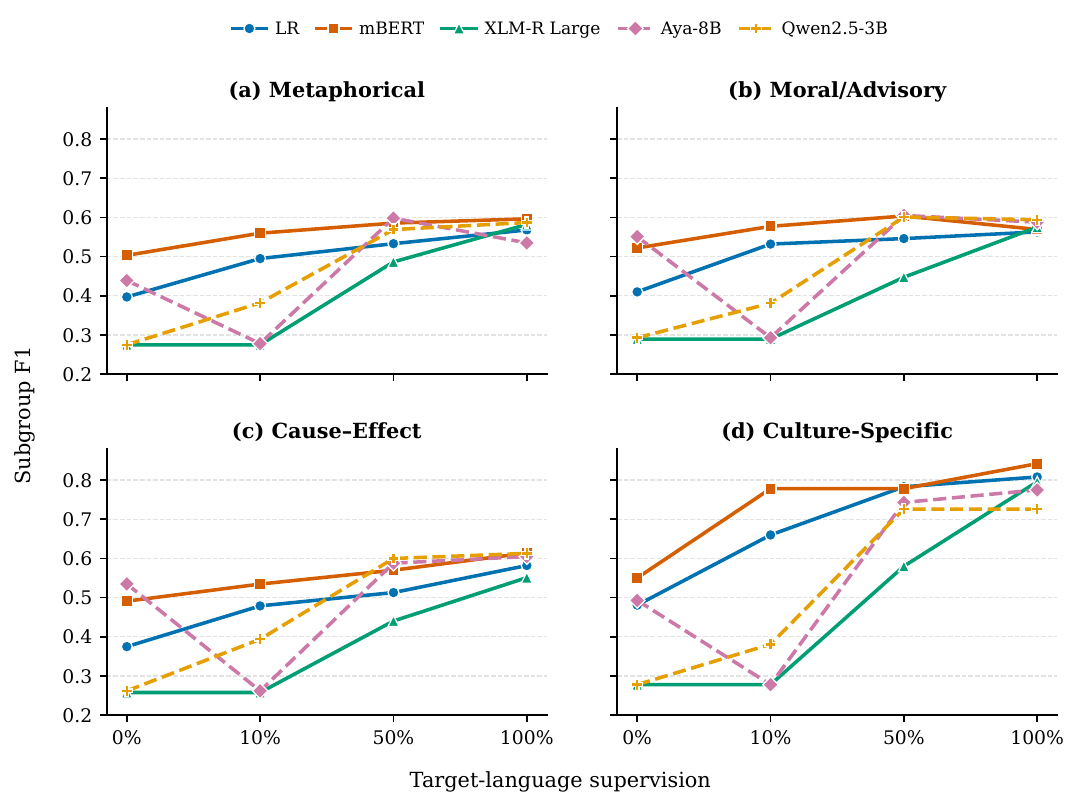}
    \caption{
    Test-set subgroup F1 across multilingual supervision levels for four fine-grained figurative properties, where 100\% denotes training on all available multilingual data.
    }
    \label{fig:finegrained-f1-by-category}
\end{figure}

\begin{table*}[t]
\scriptsize
\centering
\begin{threeparttable}

\resizebox{\textwidth}{!}{%
\begin{tabular}{lccccc}
\toprule
\textbf{Model}
& \textbf{Source core}
& \textbf{+$K$ English}
& \textbf{+$K$ multilingual}
& \(\boldsymbol{\Delta_{\mathrm{M-E}}}\)
& \textbf{95\% CI} \\
\midrule

Logistic Regression
& 0.4207
& 0.4604
& \textbf{0.5114}
& \textbf{+0.0510}
& [0.0142, 0.0856] \\

mBERT
& 0.4968
& 0.5273
& \textbf{0.5741}
& \textbf{+0.0468}
& [0.0242, 0.0690] \\

XLM-R Large
& \textbf{0.3734}
& 0.3474
& 0.3546
& +0.0072
& [$-$0.0036, 0.0171] \\

Aya-8B
& \textbf{0.4579}
& 0.4109
& 0.4412
& \textbf{+0.0304}
& [0.0075, 0.0526] \\

Qwen2.5-3B
& 0.3675
& 0.4107
& \textbf{0.4404}
& \textbf{+0.0298}
& [0.0048, 0.0553] \\

\midrule
\textbf{Mean}
& 0.4233
& 0.4313
& \textbf{0.4643}
& \textbf{+0.0330}
& --- \\

\bottomrule
\end{tabular}%
}

\caption{Matched-size English versus multilingual control on the
non-English test sets. Both conditions add $K=300$ concept-matched
examples to the same source-language core. Values are mean Macro-F1
over five seeds; $\Delta_{\mathrm{M-E}}$ is multilingual minus English,
with concept-clustered paired-bootstrap 95\% CIs.}
\label{tab:english-vs-multilingual-matched}

\end{threeparttable}
\end{table*}
We provide McNemar significant test for comparison between the models and progressive supervision levels and we find that all five models show significant differences between 0\% and 100\% multilingual supervision after Holm correction ($p_{\mathrm{Holm}} \leq .011$), with $n_{01}>n_{10}$ in every case. Thus, full multilingual supervision consistently corrects more test errors than it introduces. At 100\% supervision, only LR and Qwen2.5-3B differ significantly after Holm correction ($p_{\mathrm{Holm}}=.027$) with  all other pairwise model differences are non-significant, detail analysis in Appendix~\ref{app:validation}, Tables ~\ref{tab:mcnemar-models} and ~\ref{tab:mcnemar-supervision}.
Moreover, we provide the full table detailing training progression over each level per language in Appendix ~\ref{app:detailResults}. 
In addition, we validate on match size training and our findings show that  multilingual additions outperform English-only additions by an average of 3.3 Macro-F1 points on the non-English test sets (as shown in Table~\ref{tab:english-vs-multilingual-matched}). The difference is significant for four models; XLM-R Large shows no reliable difference because its confidence interval includes zero. Also, for per-language results, as indicated in Appendix Figure~\ref{fig:per-language-english-vs-multilingual}, show that the effect varies across languages and model architectures. The gains are most consistent for Arabic and generally positive for French, German, Russian, and Spanish. Japanese shows the weakest and most mixed transfer. 
For \textbf{RQ2}, Table~\ref{tab:controlled_instance_ablation} presents a controlled instance-selection ablation in which the multilingual supervision set is fixed to 600 examples. Overall, incorporating the complete multilingual training set provides the highest performance across all models. Our findings further show that the proposed figurative forms provide complementary supervision signals. Although each form contributes differently across model families, integrating them consistently yields the strongest multilingual performance. While \textit{Metaphorical} provides the strongest gains for mBERT and Aya, \textit{Moral/Advisory} benefits Qwen2.5-3B, and \textit{Cause-Effect} produces the best individual result for XLM-R Large. These findings suggest that different architectures exploit different meanings of figurative knowledge, whereas integrating all figurative forms provides the most robust multilingual supervision. Furthermore, Figure~\ref{fig:finegrained-f1-by-category} shows detailed performance for test-set subgroup macro F1. Among the four figurative forms, culture-specific figurative form benefits most from multilingual supervision in identifying the binary classification of figurative detection task. 
Because Culture-Specific is also
the least frequent form, we conducted an additional concept level
uncertainty and significance analysis
(Table~\ref{tab:culture-specific-uncertainty}).
Overall, all five models showed positive Culture-Specific Macro-F1 gains,
ranging from $+0.282$ to $+0.448$, and all paired
concept-clustered 95\% bootstrap intervals excluded zero.
Paired randomization tests are significant for LR, mBERT, and
XLM-R Large, but not for the other models (Aya-8B and Qwen2.5-3B). This validation 
supports the overall benefit from multilingual supervision while requiring caution for individual models because the subgroup contains only limited number of unique proverb concepts.

\begin{figure}[h]
    \centering
    \includegraphics[
        width=\columnwidth
    ]{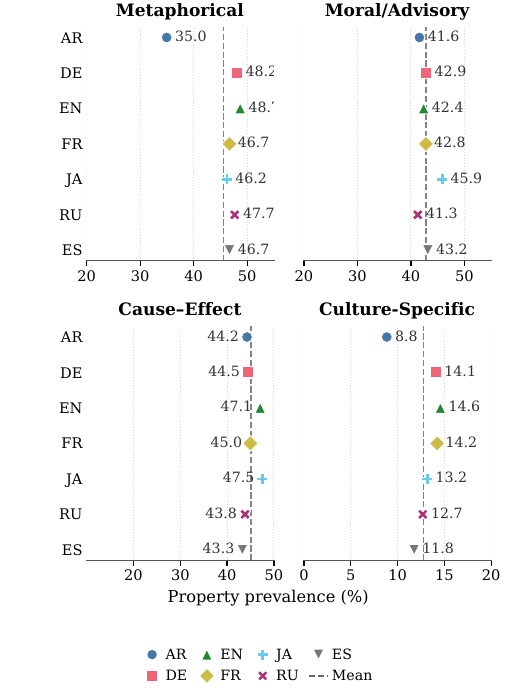}
    \caption{
    Prevalence of the four fine-grained figurative forms
    across the seven target languages. Points show language-level prevalence and dashed lines
    indicate the cross-language mean for each property.
    With multilabels percentages are calculated independently and are not
    expected to sum to 100\%.
    }
    \label{fig:finegrained-property-distribution}
\end{figure}

To answer \textbf{RQ3}, Figure~\ref{fig:finegrained-property-distribution} compares the prevalence of the four figurative forms across the seven translated languages. \textit{Moral/Advisory} and \textit{Cause-Effect} show the most stable distributions across languages Also, the \textit{Metaphorical} is consistent and show a noticible reduction in Arabic. \textit{Culture-Specific} remains the least frequent form in every language and shows greater proportional variation because of its lower prevalence.

The omnibus language-level tests reported in Appendix~\ref{app:validation} ,Table~\ref{tab:property-language-omnibus} show that none of the four figurative forms represent a significant overall language effect after Holm correction. These results indicate that the translated proverbs generally preserve the same distribution of figurative forms across languages. However, this conclusion applies at the aggregate distribution level and does not imply that every individual translated proverb retains exactly the same fine-grained labels.

\section{Discussion}
\begin{table}[t]
\centering

\scriptsize
\setlength{\tabcolsep}{2.5pt}
\resizebox{\columnwidth}{!}{%
\begin{tabular}{lccccc}
\toprule
\textbf{Model}
& \(\mathbf{F1}_{0\%}\)
& \(\mathbf{F1}_{100\%}\)
& \(\boldsymbol{\Delta}\)
& \textbf{95\% CI}
& \(\boldsymbol{p}\) \\
\midrule
LR
& .481 & .808 & +.327
& [.167, .441] & \textbf{.024} \\

mBERT
& .550 & \textbf{.842} & +.292
& [.155, .441] & \textbf{.002} \\

XLM-R Large
& .468 & .815 & +.348
& [.141, .491] & \textbf{.013} \\

Aya-8B
& .493 & .775 & +.282
& [.024, .557] & .082 \\

Qwen2.5-3B
& .278 & .726 & \textbf{+.448}
& [.151, .653] & .090 \\
\bottomrule
\end{tabular}%
}
\caption{Culture-Specific performance and uncertainty. Results use 109
test instances. CIs are paired
concept-clustered bootstrap intervals. We use $p$-values 
significance at $\alpha=0.05$.}
\label{tab:culture-specific-uncertainty}
\end{table}
\paragraph{Different figurative forms respond differently to multilingual supervision, even though their overall prevalence is preserved across languages.}~Figure~\ref{tab:controlled_instance_ablation} and Figure~\ref{fig:finegrained_jaccard} together provide an insight for an important characteristic of multilingual figurative-language learning in proverbs. Although the four figurative forms are preserved across translated proverbs (as shown in Figure ~\ref{fig:finegrained-property-distribution}) yet, they do not benefit equally from multilingual supervision (Figure~\ref{tab:controlled_instance_ablation}). Particularly, Culture-Specific has the largest performance gains despite being the least prevalent figurative form and the one with the lowest overlap with the remaining forms. As in the dataset, Culture-Specific accounts for 705 of the 4,777 training instances (14.8\%), compared with 109 of the 1,282 test instances (8.5\%). In contrast, Metaphorical, Moral/Advisory, and Cause-Effect show substantially greater overlap with one another, suggesting that these forms share complementary semantic cues that can be learned jointly. These findings indicate that multilingual supervision does not simply reinforce the existing semantic distribution of the data. Rather, it differentially enhances the learning of figurative forms according to their semantic distinctiveness. 
For example, the proverb \textit{Curiosity killed the cat} is simultaneously metaphorical, advisory, and causal: the cat imagery conveys an indirect warning, while the proverb links excessive curiosity to a harmful
consequence. As its multilingual realizations preserve this shared structure even when the wording changes, such as the French \textit{La curiosité est un vilain défaut}. This observation highlights the importance of moving beyond treating figurative language in proverbs as a single phenomenon and motivates future multilingual evaluation frameworks that explicitly distinguish between complementary forms of figurative meanings related to the context of evaluation in proverbs. Prior work has shown that proverb translation quality varies across cultural and linguistic settings and that traditional translation metrics may not adequately capture cultural or figurative equivalence \citep{Wang2025-at}. Similarly, multilingual models may recognize or translate proverbs while still exhibiting a culture gap in contextual
reasoning \citep{Liu2024-ab}. On the same line, our findings therefore distinguish between \emph{distributional preservation},the comparable prevalence of figurative forms across languages in proverbs, and \emph{instance-level preservation}, which requires direct concept-level evaluation of whether each translation retains the same figurative and cultural interpretation.

\paragraph{Culture-specific meaning and the limits of categorical preservation.} In this study the analysis of Culture-Specific is particularly informative because its low overlap with the other forms suggests that it captures a comparatively distinct source of figurative knowledge. A qualitative examination in Appendix~\ref{tab:culture-specific-qualitative-example} further shows that a shared Culture-Specific label does not guarantee cultural equivalence across translations. For example, the concept expressed by \textit{No sweet without sweat} is interpreted through different culturally familiar images across languages, including diligence and reward in German and labor, and food in French. Although these inferences preserve the general conclusion that effort precedes reward, yest still the cultural symbols and knowledge differ. Thus, categorical consistency can demonstrate the presence of cultural grounding, but cannot determine whether the underlying cultural meaning is preserved or localized. This motivate future concept-level analysis that compares aligned translations beyond label consistency and examines whether the observed gains for Culture-Specific forms are driven by multilingual lexical diversity, culturally distinct, or their relative independence from other figurative forms.

\section{Conclusion}
This study examines the effectiveness of multilingual supervision for figurative language identification in multilingual proverbs. Multilingual additions generally improve non-English transfer beyond matched English additions, although effects vary across models and languages. The analysis provides a detailed examination of concept-level complementary figurative forms in proverbs. Culture-Specific shows the largest gains despite its low prevalence, motivating future multilingual benchmarks that explicitly model culturally grounded figurative meaning. These findings open the door to a new generation of multilingual figurative-language benchmarks in proverbs that move beyond metaphor-centric taxonomies by explicitly modeling complementary figurative forms, particularly culture-specific meaning.

\subsection*{Limitations}
Although the multilingual proverb dataset is labeled through weak supervision rather than manual annotation, we mitigate this limitation by grounding the silver labels in a human-validated seed set, selecting the best-performing weak-supervision models through held-out evaluation, and validating the resulting multilingual analyses using statistical significance tests. Our experiments focus on multilingual proverb translations by providing a controlled setting for studying figurative language transfer. Although, the findings may not directly generalize to other figurative phenomena such as sarcasm, irony, or humor. Moreover, our analysis characterize the transferability of complementary figurative forms at the distributional level. Yet, it does not establish whether individual translated proverbs preserve identical figurative or cultural interpretations. Addressing concept-level preservation across multilingual translations remains an important direction for future work.

\subsection*{Ethical considerations}
Human annotations were collected only for the manually validated subset following predefined annotation guidelineswith no personally identifiable information collected. We provide the detail annotation process recrutment and guideline in appendix~\ref{app:annotationValidation}. The resulting weakly supervised labels are intended to support large-scale multilingual classification of figurative labels analysis and should not be regarded as human gold annotations.

\bibliography{custom,paperpile.bib,figritiveref.bib}

\appendix

\begin{table}[t]
\centering
\small
\begin{tabular}{lc}
\toprule
\textbf{Label} & \textbf{Error Rate} $\downarrow$ \\
\midrule
Metaphorical      & 0.369 \\
Moral/Advisory    & 0.430 \\
Cause--Effect     & 0.376 \\
Culture-Specific  & \textbf{0.181} \\
None/Literal      & 0.295 \\
\midrule
\textbf{Overall Hamming error} & \textbf{0.330} \\
\textbf{Sample Jaccard} $\uparrow$ & \textbf{0.439} \\
\bottomrule
\end{tabular}
\caption{Manual validation of weakly supervised silver labels over 149 matched held-out proverb concepts. Per-label values report disagreement rates.}
\label{tab:silver-label-validation}
\end{table}

\section{Human Validation and Structural Annotation}
\label{app:annotationValidation}
To provide a controlled validation layer beyond the weakly supervised labels, we constructed a mutually aligned subset of 83 proverb concepts that were available in all seven target languages: Arabic, English, French, German, Russian, Japanese, and Spanish. This subset was selected to enable direct multilingual comparison under identical proverb concepts, reducing confounds caused by uneven translation coverage across languages. Each proverb was labeled using a multiple labels with four possible figurative meanings forms labels: Metaphorical, Moral/Advisory, Cause–Effect, Culture-Specific, and None.

\begin{table*}[t]
\centering
\scriptsize
\begin{tabular}{p{0.18\textwidth} p{0.36\textwidth} p{0.32\textwidth}}
\hline
\textbf{Annotation Component} & \textbf{Description} & \textbf{Example and Rule} \\
\hline
Task objective & Annotators classify each proverb according to its dominant semantic or figurative structure. The goal is to capture how the proverb conveys meaning. & Annotators focus on intended meaning rather than grammar, translation quality, or personal agreement. \\
\hline
Input & Each annotation item contains a proverb and its corresponding source identifier. & \texttt{source\_id} + proverb text. \\
\hline
Output & A single-select label is assigned to each proverb. & Annotators select multiple related forms/labels. \\
\hline
Metaphorical & The proverb conveys meaning indirectly through symbolic imagery, analogy, animals, objects, nature, or figurative comparison. & ``The early bird catches the worm''  bird imagery symbolizes initiative. \\
\hline
Moral / Advisory & The proverb mainly provides advice, ethical guidance, warning, or behavioral instruction. & ``Honesty is the best policy''   ethical guidance. \\
\hline
Cause-Effect / Consequence & The proverb links actions, decisions, or conditions with consequences or outcomes. & ``You reap what you sow''   actions lead to consequences. \\
\hline
Culture-Specific Symbolic & The proverb depends on local culture, religion, historical references, traditions, or culturally specific symbolism. & Proverbs involving tribal customs, religious symbolism, or historically grounded expressions. \\
\hline
None & The proverb is treated as literal when its meaning can be understood directly and no symbolic interpretation is required. & if interpreted as direct meaning. \\
\hline
Decision rule & If multiple categories appear possible, annotators choose tall representative labels. & overlap allowed between labels as one prverb might carry multiple forms at the same time \\
\hline
\end{tabular}
\caption{Annotation guideline for proverb figurative structure classification. Annotators assigned one or more of the four figurative-form labels. The first four labels , Metaphorical, Moral/Advisory, Cause and Effect, and Culture-Specific Symbolic, were treated as figurative structure labels, while \textit{None} was treated as the literal class. Thus, the annotation schema supports both fine-grained figurative structure analysis and binary figurative-versus-literal evaluation.}
\label{tab:proverb_annotation_guidelines}
\end{table*}

Two annotators from authors independently annotated a common subset of 83 proverb concepts following the guidelines described in Table~\ref{tab:proverb_annotation_guidelines} with third annotator acted as reviewer. Table~\ref{tab:annotation_agreement} shows the inter-annotator agreement for the binary figurative-versus-literal classification achieved a Cohen's $\kappa$ of 0.626 (81.9\% raw agreement), indicating substantial agreement. For the multi-label figurative properties, Cohen's $\kappa$ ranged from 0.423 (Cause andEffect) to 0.622 (None/Literal), with the highest raw agreement observed for Culture-Specific Symbolic (92.8\%). Agreement over the complete multilabel annotation vectors yielded an exact-match agreement of 55.4\%, a sample-wise Jaccard similarity of 67.9\%, and a Hamming loss of 16.1\%, indicating that although annotators occasionally differed on individual forms of figurative forms, they generally identified highly overlapping sets of figurative characteristics for each proverb.

\begin{table}[t]
\centering
\scriptsize
\begin{tabular}{lcc}
\hline
\textbf{Label} & \textbf{Cohen's $\kappa$} & \textbf{Agree(\%)} \\
\hline
Figurative vs. Literal & 0.626 & 81.9 \\
\hline
Metaphorical & 0.595 & 79.5 \\
Moral / Advisory & 0.576 & 85.5 \\
Cause-Effect & 0.423 & 79.5 \\
Culture-Specific Symbolic & 0.533 & 92.8 \\
None (Literal) & 0.622 & 81.9 \\
\hline
Exact Match (Multilabel) & -- & 55.4 \\
Sample-wise Jaccard Similarity & -- & 67.9 \\
Hamming Loss$\downarrow$ & -- & 16.1 \\
\hline
\end{tabular}
\caption{Inter-annotator agreement for the proverb annotation task on the subset of 83 commonly annotated proverb concepts. Cohen's $\kappa$ is reported for the binary figurative-versus-literal task and for each multilabel property.}
\label{tab:annotation_agreement}
\end{table}

\paragraph{Weak supervision Figurative binary Labeling}

The combined ground-truth dataset was partitioned into training (80\%), validation (10\%), and test (10\%) sets using a two-stage stratified split. First, 20\% of the data was held out from the training set and subsequently divided equally into validation and test partitions. Stratification was performed on the binary figurative-language label at both stages to preserve the overall class distribution across all splits. The final dataset comprises 325,298 instances, including 260,238 training, 32,530 validation, and 32,530 test examples. Overall, the dataset contains 139,002 figurative (42.7\%) and 186,296 literal (57.3\%) instances, corresponding to a class balance ratio of 0.746. All partitions were generated using a fixed random seed of 42, and data integrity was verified by confirming that the total number of instances remained unchanged after splitting. As shown in Table~\ref{tab:weak-supervision-mcnemar} and Table~\ref{tab:weak-supervision-binary-performance}, We compare the performance of three baseline models along with XLM-RoBERTa Base. Based on the performance and McNemar evaluation we selected was selected XLM-RoBERTa Base. The fine-tuned model from the second experiment was applied to all unique source proverbs in the multilingual dataset to generate silver labels. The generated labels were then propagated to all translations sharing the same source proverb identifier, ensuring that conceptually equivalent proverbs received consistent labels regardless of language. Since weak supervision inherently produces noisy labels, a small number of targeted manual corrections were applied to verifiable Arabic and English proverbs to mitigate the impact of obvious errors before passing the final labels to downstream models.

\begin{figure*}[t]
    \centering
    \includegraphics[width=\textwidth]{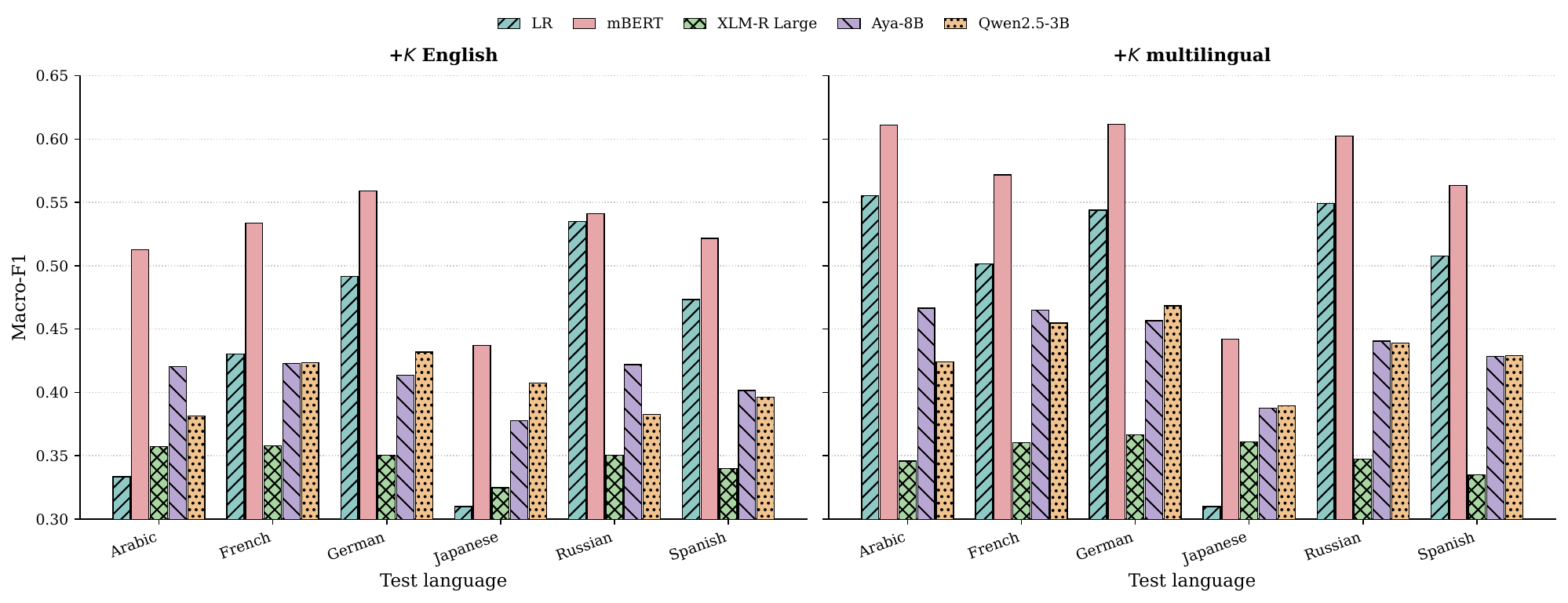}
    \caption{Validation Per-language Macro-F1 for the matched-size English and multilingual supervision conditions. Both conditions augment the same source-language core with $K=300$ concept-matched examples. Multilingual addition is balanced across the six non-English languages (50 examples per language). Results averaged over five seeds and evaluated on the same held-out nonEnglish test sets.}
    \label{fig:per-language-english-vs-multilingual}
\end{figure*}

\paragraph{Weak Supervision Fine Grain multilabeling}

For multi labeling fine grain figurative forms, we compare multiple baseline models as reported in Table~\ref{tab:finegrained_results} and we find that
E5 embeddings with Logistic Regression achieved the highest Macro-F1
($0.3640$), outperforming the Random Prevalence Baseline
($0.2765$) and the Majority Baseline ($0.1772$). This indicates that the model learns meaningful patterns beyond label-frequency statistics. Although the Majority Baseline achieved the highest sample-level Jaccard, this primarily reflects its tendency to predict dominant label combinations rather than balanced performance across the four fine-grained labels. We select E5 with Logistic Regression for confidence-aware silver-label generation.

\paragraph{Error Analysis of the Weak Supervision Labels}

Three of the authors further manually validated the silver labels for 149 held out proverb concepts, matching instances by \texttt{sourceid}.
As shown in Table~\ref{tab:silver-label-validation}, agreement then assessed independently for the five fine-grained labels which show a low Hamming error rate of 0.330.
The manual and silver label sets achieved a moderate sample level Jaccard similarity = 0.439.
Per-label error rates ranged from 0.181 for Culture-Specific to 0.430 for Moral/Advisory, indicating variation in silver-label reliability across categories.
\begin{table*}[t]
\centering
\scriptsize
\setlength{\tabcolsep}{3.2pt}
\renewcommand{\arraystretch}{1.08}

\begin{tabular}{lccccc|ccccc|ccccc}
\toprule
& \multicolumn{5}{c}{\textbf{LR}}
& \multicolumn{5}{c}{\textbf{mBERT}}
& \multicolumn{5}{c}{\textbf{XLM-R Large}} \\
\cmidrule(lr){2-6}
\cmidrule(lr){7-11}
\cmidrule(lr){12-16}

\textbf{Testset}
& \textbf{0\%}
& \textbf{10\%}
& \textbf{50\%}
& \textbf{100\%}
& \textbf{$\Delta$}
& \textbf{0\%}
& \textbf{10\%}
& \textbf{50\%}
& \textbf{100\%}
& \textbf{$\Delta$}
& \textbf{0\%}
& \textbf{10\%}
& \textbf{50\%}
& \textbf{100\%}
& \textbf{$\Delta$} \\
\midrule

Metaphor
& 0.397 & 0.495 & 0.533 & 0.568 & +0.171
& 0.5032&	0.5598&	0.5855&	0.5964&	+0.0932
& 0.2749
&0.2749
&0.4865
&0.5820
&+0.307 \\

Moral
& 0.410 & 0.532 & 0.546 & 0.563 & +0.153
& 0.5221&	0.5772&	0.6037	&0.5692	&+0.0471
& 0.2896
&0.2896
&0.4473
&0.5751
&+0.2855 \\

Cause
& 0.375 & 0.479 & 0.513 & 0.582 & +0.207
& 0.4910&	0.5347&	0.5701&	0.6141&	+0.1231
& 0.2580
&0.2580
&0.4404
&0.5511
&+0.293 \\

Culture
& 0.481 & 0.660 & 0.783 & \textbf{0.808} & \textbf{+0.327}
& 0.5503&	0.7782&	0.7782	&\textbf{0.8420}&	\textbf{+0.2917}
&0.2781
&0.2781
&0.5802
&0.7957
&+0.5176 \\

None
& 0.456 & 0.406 & 0.396 & 0.376 & -0.080
& 0.4088	&0.3729&	0.3717&	0.3570&	-0.0518
&0.4078&
0.4078&
0.4296&
0.5126&
+0.1048\\

\midrule
\textbf{All}
& 0.463 & 0.540 & 0.561 & 0.572 & +0.108
& 0.5161&	0.5911&	0.6108&	\textbf{0.6072}&	+0.0910
& 0.3402 & 0.3402 & 0.4872 & 0.5811
& +0.240 \\

\bottomrule
\end{tabular}
\caption{Binary figurative-identification performance across fine-grained diagnostic subsets for Logistic Regression (LR), mBERT, XLM-R Base, and XLM-R Large under progressively increasing multilingual supervision. Each cell reports subgroup F1, and $\Delta$ denotes the absolute gain from 0\% to 100\% training supervision. Fine-grained subsets are non-exclusive because a proverb may carry multiple labels.}
\label{tab:finegrained_supervision_four_models}
\end{table*}

\begin{table*}[t]
\centering
\scriptsize
\setlength{\tabcolsep}{3.5pt}
\renewcommand{\arraystretch}{1.08}

\begin{tabular}{lccccc|ccccc}
\toprule
&
\multicolumn{5}{c}{\textbf{Aya-8B}}
&
\multicolumn{5}{c}{\textbf{Qwen2.5-3B}}\\

\cmidrule(lr){2-6}
\cmidrule(lr){7-11}

\textbf{Test Subset}
&
\textbf{0\%}
&
\textbf{10\%}
&
\textbf{50\%}
&
\textbf{100\%}
&
\textbf{$\Delta$}
&
\textbf{0\%}
&
\textbf{10\%}
&
\textbf{50\%}
&
\textbf{100\%}
&
\textbf{$\Delta$}
\\

\midrule

Metaphorical
&
0.439
&
0.278
&
0.598
&
0.535
&
+0.095
&
0.275
&
0.382
&
0.569
&
0.587
&
+0.312
\\

Moral/Advisory
&
0.551
&
0.293
&
0.605
&
0.588
&
+0.037
&
0.293
&
0.381
&
0.601
&
0.594
&
+0.301
\\

Cause-Effect
&
0.535
&
0.262
&
0.588
&
0.605
&
+0.070
&
0.262
&
0.394
&
0.600
&
0.613
&
+0.351
\\

Culture-Specific
&
0.493
&
0.278
&
0.743
&
\textbf{0.775}
&
\textbf{+0.282}
&
0.278
&
0.381
&
0.726
&
\textbf{0.726}
&
\textbf{+0.448}
\\

Literal/None
&
0.423
&
0.500
&
0.429
&
0.328
&
-0.096
&
0.500
&
0.003
&
0.394
&
0.370
&
-0.130
\\

\midrule

\textbf{All}
&
0.524
&
0.342
&
\textbf{0.616}
&
0.597
&
+0.073
&
0.342
&
0.329
&
\textbf{0.624}
&
0.620
&
+0.278
\\

\bottomrule
\end{tabular}

\caption{Binary figurative-identification performance across fine-grained diagnostic subsets for multilingual instruction-tuned LLMs under progressively increasing multilingual supervision. Each cell reports Macro-F1, and $\Delta$ denotes the absolute gain from 0\% to 100\% supervision. Fine-grained subsets are non-exclusive because a proverb may carry multiple labels.}

\label{tab:finegrained_supervision_llms}
\end{table*}
\section{Per-Language and Progression Training Level Detail Results}
\label{app:detailResults}
Table~\ref{tab:finegrained_supervision_four_models} and ~\ref{tab:finegrained_supervision_llms} show detailed performance for binary figurative-identification performance across fine-grained diagnostic subsets. Also, Table~\ref{tab:per-language-all-levels} provides the complete
per-language results across all supervision levels and models.
Performance varies considerably by language, model, and supervision
level, with the strongest model often changing as additional
multilingual data are introduced.


\begin{table}[t]
\centering
\scriptsize
\setlength{\tabcolsep}{4.5pt}
\renewcommand{\arraystretch}{1.08}
\begin{tabular}{lccc}
\toprule
\textbf{Model}  & \textbf{Macro-F1} & \textbf{(0) F1} & \textbf{(1) F1} \\
\midrule
XLM-R Base & \textbf{0.9423} & 0.9507 & 0.9339 \\
TF-IDF + Linear SVM & 0.9123 & 0.9277 & 0.8968 \\
TF-IDF + LR & 0.9095  & 0.9237 & 0.8889 \\
TF-IDF + Multinomial NB  & 0.8508 & 0.8825 & 0.8190 \\
\bottomrule
\end{tabular}
\caption{Performance of the binary weak-supervision, (0) denotes Literal label, and (1) denotes figurative label.}
\label{tab:weak-supervision-binary-performance}
\end{table}
\begin{table}[t]
\centering
\scriptsize
\setlength{\tabcolsep}{4pt}
\renewcommand{\arraystretch}{1.08}
\begin{tabular}{lrrrr}
\toprule
\textbf{Comparison} & \textbf{$n_{10}$} & \textbf{$n_{01}$} & \textbf{Stat} & \textbf{$p_{\mathrm{Holm}}$} \\
\midrule
 MNB vs.\ XLM-R Base & 885 & 3684 & 1713.461 & \textbf{$<.001$} \\
LR vs.\ XLM-R Base & 799 & 1906 & 452.213 & \textbf{$<.001$} \\
 SVM vs.\ XLM-R Base & 863 & 1792 & 324.363 & \textbf{$<.001$} \\
\bottomrule
\end{tabular}
\caption{Pairwise McNemar comparisons between XLM-R Base and TF-IDF classical weak-supervision baselines.  $n_{10}$ denotes instances correctly classified by the first model but not the second and $n_{01}$ denotes the reverse.  $p$-values show Holm-adjusted across pairwise comparisons.}
\label{tab:weak-supervision-mcnemar}
\end{table}

\begin{table}[t]
\centering
\scriptsize
\setlength{\tabcolsep}{5pt}
\begin{tabular}{lcc}
\toprule
\textbf{Model} &
\textbf{Macro-F1 $\uparrow$} &
\textbf{Sample Jaccard $\uparrow$} \\
\midrule
E5 + LR
& \textbf{0.3640 $\pm$ 0.1045}
& 0.2667 $\pm$ 0.0862 \\

E5 + Classifier Chain
& 0.3112 $\pm$ 0.1021
& 0.1336 $\pm$ 0.0345 \\

Random Baseline
& 0.2765 $\pm$ 0.0086
& 0.2062 $\pm$ 0.0100 \\

E5 + Ridge
& 0.2702 $\pm$ 0.0757
& 0.2898 $\pm$ 0.0842 \\

Majority Baseline
& 0.1772 $\pm$ 0.0035
& \textbf{0.3386 $\pm$ 0.0267} \\
\bottomrule
\end{tabular}
\caption{Weak-supervision for multi-label fine grain labels models comparison. Repeated $5{\times}5$ multi-label cross-validation results.sample-level Jaccard evaluates the overlap between the predicted and gold label sets for each proverb.}
\label{tab:finegrained_results}
\end{table}
\begin{table*}[t]
\centering
\caption{
Per-language Macro-F1 across multilingual supervision levels.
Bold values indicate the highest-performing model for each language
and supervision level.
}
\label{tab:per-language-all-levels}
\setlength{\tabcolsep}{7pt}
\renewcommand{\arraystretch}{1.08}
\small
\begin{tabular}{llccccc}
\toprule
\textbf{Language}
& \textbf{Supervision}
& \textbf{LR}
& \textbf{mBERT}
& \textbf{XLM-R Large}
& \textbf{Aya-8B}
& \textbf{Qwen2.5-3B} \\
\midrule

\multirow{4}{*}{Arabic}
& 0\%   & 0.333 & \textbf{0.452} & 0.333 & 0.399 & 0.333 \\
& 10\%  & 0.376 & \textbf{0.614} & 0.333 & 0.333 & 0.333 \\
& 50\%  & 0.440 & 0.636 & 0.333 & \textbf{0.751} & 0.582 \\
& 100\% & 0.500 & 0.615 & 0.333 & 0.617 & \textbf{0.674} \\
\midrule

\multirow{4}{*}{English}
& 0\%   & 0.530 & 0.503 & 0.358 & \textbf{0.587} & 0.367 \\
& 10\%  & 0.535 & \textbf{0.610} & 0.358 & 0.367 & 0.322 \\
& 50\%  & 0.531 & 0.594 & 0.514 & 0.603 & \textbf{0.625} \\
& 100\% & 0.575 & 0.572 & 0.591 & 0.582 & \textbf{0.604} \\
\midrule

\multirow{4}{*}{French}
& 0\%   & 0.353 & \textbf{0.550} & 0.355 & 0.544 & 0.355 \\
& 10\%  & 0.577 & \textbf{0.586} & 0.355 & 0.355 & 0.311 \\
& 50\%  & 0.530 & 0.611 & 0.533 & \textbf{0.639} & 0.629 \\
& 100\% & 0.531 & 0.639 & 0.600 & 0.621 & \textbf{0.661} \\
\midrule

\multirow{4}{*}{German}
& 0\%   & \textbf{0.497} & 0.465 & 0.347 & 0.482 & 0.347 \\
& 10\%  & 0.540 & \textbf{0.629} & 0.347 & 0.347 & 0.319 \\
& 50\%  & 0.596 & 0.638 & 0.440 & 0.639 & \textbf{0.641} \\
& 100\% & 0.605 & 0.628 & 0.565 & 0.584 & \textbf{0.629} \\
\midrule

\multirow{4}{*}{Japanese}
& 0\%   & 0.310 & 0.488 & 0.310 & \textbf{0.514} & 0.310 \\
& 10\%  & 0.310 & \textbf{0.409} & 0.310 & 0.310 & 0.350 \\
& 50\%  & 0.350 & 0.476 & 0.310 & 0.479 & \textbf{0.502} \\
& 100\% & 0.449 & 0.527 & \textbf{0.597} & 0.513 & 0.493 \\
\midrule

\multirow{4}{*}{Russian}
& 0\%   & \textbf{0.535} & 0.510 & 0.342 & 0.481 & 0.342 \\
& 10\%  & 0.569 & \textbf{0.592} & 0.342 & 0.342 & 0.325 \\
& 50\%  & 0.598 & 0.650 & 0.502 & 0.643 & \textbf{0.665} \\
& 100\% & 0.591 & 0.621 & 0.565 & 0.641 & \textbf{0.660} \\
\midrule

\multirow{4}{*}{Spanish}
& 0\%   & 0.328 & 0.418 & 0.312 & \textbf{0.480} & 0.312 \\
& 10\%  & 0.575 & \textbf{0.645} & 0.312 & 0.312 & 0.353 \\
& 50\%  & \textbf{0.654} & 0.649 & 0.531 & 0.627 & 0.646 \\
& 100\% & 0.616 & \textbf{0.631} & 0.566 & 0.609 & 0.609 \\

\bottomrule
\end{tabular}
\end{table*}
\section{Models setting}
\label{app:models_settings}
\textbf{ Logistic Regression} The TF-IDF vectorizer was configured with a unigram and bigram range (\texttt{ngram\_range=(1,2)}), a minimum document frequency of 2, and a maximum of 20,000 features. The Logistic Regression classifier was trained with a maximum of 1,000 iterations.
\textbf{Multilingual BERT (mBERT)} The model was fine-tuned for 2 epochs with a batch size of 16, a learning rate of 2e-5, and a weight decay of 0.01. A linear learning rate scheduler with a 10\% warmup period was applied to stabilize early training.
\textbf{XLM-RoBERTa}Large the modle trained for 3 epochs using the AdamW optimizer. XLM-R Base was fine-tuned with a learning rate of 2e-5 and a batch size of 32, while XLM-R Large was fine-tuned with a learning rate of 5e-6 and the same batch size. \textbf{Aya Expanse 8B} We fine-tuned Aya Expanse 8B using QLoRA with 4-bit NF4 quantization, LoRA adapters ($r=16$, $\alpha=32$, dropout $=0.05$), one training epoch, a learning rate of $2\times10^{-4}$, cosine scheduling, and gradient checkpointing. The effective batch size was 16 (2 × 8 gradient accumulation). \textbf{Qwen2.5-3B} We fine-tuned Qwen2.5-3B-Instruct using the same QLoRA configuration and optimization settings. The effective batch size was 16 (4 × 4 gradient accumulation).

\section{Validation of the analysis}
\label{app:validation}
\paragraph{Omnibus tests of cross-language differences in the prevalence of fine-grained forms of figurative forms} To validate the multilingual consistency of the analysis. As shown in Table~\ref{tab:property-language-omnibus} we fitted separate logistic GEE models for each semantic property, clustering observations by proverb concept. After Holm correction across the four omnibus tests, none of the language effects remained statistically significant. This indicates that the overall prevalence of the fine-grained properties was broadly stable across languages, supporting their use as cross-linguistic semantic dimensions while allowing for limited descriptive variation.

\paragraph{McNemar significance analysis of models performance.}
As shown in Tables ~\ref{tab:mcnemar-models} and ~\ref{tab:mcnemar-supervision} all five models show significant differences between 0\% and 100\% multilingual supervision after Holm correction ($p_{\mathrm{Holm}} \leq .011$), with $n_{01}>n_{10}$ in every case. Thus, full multilingual supervision consistently corrects more test errors than it introduces. At 100\% supervision, only LR and Qwen2.5-3B differ significantly after Holm correction ($p_{\mathrm{Holm}}=.027$); all other pairwise model differences are non-significant.

\begin{table}[t]
\centering
\scriptsize
\setlength{\tabcolsep}{6pt}
\renewcommand{\arraystretch}{1.08}
\begin{tabular}{lrrrr}
\toprule
\textbf{Comparison} &
$\mathbf{n_{10}}$ &
$\mathbf{n_{01}}$ &
$\boldsymbol{\chi^2}$ &
$\mathbf{p_{\text{Holm}}}$ \\
\midrule
LR: 0\% vs 100\%          & 230 & 291 &  6.910 & \textbf{0.011} \\
mBERT: 0\% vs 100\%       & 215 & 312 & 17.488 & \textbf{$<$.001} \\
XLM-R Large: 0\% vs 100\% & 248 & 334 & 12.414 & \textbf{0.001} \\
Aya-8B: 0\% vs 100\%      & 293 & 365 &  7.661 & \textbf{0.011} \\
Qwen2.5-3B: 0\% vs 100\%  & 272 & 405 & 25.737 & \textbf{$<$.001} \\
\bottomrule
\end{tabular}
\caption{McNemar tests comparing source-only and full multilingual supervision. Holm-adjusted significant values are shown in bold.}
\label{tab:mcnemar-supervision}
\end{table}
\begin{table}[t]
\centering
\scriptsize
\setlength{\tabcolsep}{5pt}
\renewcommand{\arraystretch}{1.08}
\begin{tabular}{lrrrr}
\toprule
\textbf{Comparison} &
$\mathbf{n_{10}}$ &
$\mathbf{n_{01}}$ &
$\boldsymbol{\chi^2}$ &
$\mathbf{p_{\text{Holm}}}$ \\
\midrule
LR vs mBERT              & 147 & 192 & 5.711 & 0.152 \\
LR vs XLM-R        & 257 & 270 & 0.273 & 1.000 \\
LR vs Aya             & 173 & 210 & 3.384 & 0.461 \\
LR vs Qwen         & 170 & 231 & 8.978 & \textbf{0.027} \\
mBERT vs XLM-R      & 246 & 214 & 2.089 & 0.742 \\
mBERT vs Aya          & 118 & 110 & 0.215 & 1.000 \\
mBERT vs Qwen     & 120 & 136 & 0.879 & 1.000 \\
XLM-R vs Aya   & 217 & 241 & 1.155 & 1.000 \\
XLM-R vs Qwen& 190 & 238 & 5.161 & 0.185 \\
Aya vs Qwen     &  99 & 123 & 2.383 & 0.736 \\
\bottomrule
\end{tabular}
\caption{Pairwise McNemar tests among models under full multilingual supervision. Holm-adjusted significant values are shown in bold.}
\label{tab:mcnemar-models}
\end{table}


\begin{table}[t]
\centering
\small
\begin{threeparttable}
\caption{
Omnibus tests of cross-language differences in the prevalence of
fine-grained forms of figurative forms.
}
\label{tab:property-language-omnibus}
\begin{tabular}{lrrrr}
\toprule
\textbf{Property}
& \(\boldsymbol{\chi^2}\)
& \textbf{\(df\)}
& \textbf{ \(p\)}
& \textbf{Holm \(p\)} \\
\midrule
Metaphorical     & 14.65 & 6 & .023 & .070 \\
Moral/Advisory   &  8.12 & 6 & .229 & .229 \\
Cause-Effect    & 16.08 & 6 & .013 & .053 \\
Culture-Specific & 12.49 & 6 & .052 & .104 \\
\bottomrule
\end{tabular}
\begin{tablenotes}[flushleft]
\footnotesize
\item Separate logistic generalized estimating equation models were
fitted for each binary property, with language as a categorical
predictor and robust standard errors clustered by proverb concept.
\end{tablenotes}
\end{threeparttable}
\end{table}

\begin{table}[t]
\centering
\scriptsize
\setlength{\tabcolsep}{4pt}
\renewcommand{\arraystretch}{1.12}

\begin{tabular}{
    p{0.08\textwidth}    
    p{0.10\textwidth}
    p{0.20\textwidth}
}
\toprule
\textbf{Trans.\ ID 
(Langt)} &
\textbf{Aligned proverb} &
\textbf{Approximate English gloss} \\
\midrule

945642 (DE)  &
\textit{Ohne Fleiß kein Preis.} &
No prize without diligence; replaces sweat with diligence. \\

18370 (EN)  &
\textit{No pain, no gain.} &
Uses pain and gain to convey the same general lesson. \\

2991444 (FR)  &
\textit{C'est par le labeur que l'on fait son beurre.} &
Through labor one makes one's butter; uses work and food imagery. \\

6361745 (ES)  &
\textit{No hay premio sin esfuerzo.} &
There is no reward without effort. \\

\bottomrule
\end{tabular}

\caption{Aligned multilingual realizations of the Culture-Specific proverb
concept \textit{No sweet without sweat}. The realizations saved
a broadly shared effort and reward relation, yet they use different culturally
familiar images. }
\label{tab:culture-specific-qualitative-example}
\end{table}

\end{document}